\documentclass[11pt,a4paper]{article}
\usepackage[hyperref]{emnlp2020}
\usepackage{times}
\usepackage{latexsym}
\usepackage{url}
\usepackage{multirow}
\usepackage{graphicx}
\usepackage{amsmath}
\usepackage{amssymb}
\usepackage{enumitem}
\usepackage{quoting}
\usepackage{makecell}

\usepackage{microtype}

\aclfinalcopy
\def\aclpaperid{911}

\title{LUKE: Deep Contextualized Entity Representations with\\ Entity-aware Self-attention}
\author{
    Ikuya Yamada$^{1,2}$\\
    {\small \texttt{ikuya@ousia.jp}}
    \And
    Akari Asai$^3$\\
    {\small \texttt{akari@cs.washington.edu}}
    \And
    Hiroyuki Shindo$^{4,2}$\\
    {\small \texttt{shindo@is.naist.jp}}
    \AND
    Hideaki Takeda$^5$\\
    {\small \texttt{takeda@nii.ac.jp}}
    \And
    Yuji Matsumoto$^{2}$\\
    {\small \texttt{matsu@is.naist.jp}}
    \AND
    \begin{minipage}{\textwidth}
        \begin{center}
            \fontsize{11.5}{14}\selectfont
            \textnormal{$^1$Studio Ousia\,\,\,$^2$RIKEN AIP\,\,\,$^3$University of Washington}\\
            \textnormal{$^4$Nara Institute of Science and Technology\,\,\,$^5$National Institute of Informatics} \end{center}
    \end{minipage}
}
\date{}

\begin{document}
\maketitle
\begin{abstract}
    Entity representations are useful in natural language tasks involving entities.
    In this paper, we propose new pretrained contextualized representations of words and entities based on the bidirectional transformer \cite{NIPS2017_7181}.
    The proposed model treats words and entities in a given text as independent tokens, and outputs contextualized representations of them.
    Our model is trained using a new pretraining task based on the masked language model of BERT \cite{devlin2018bert}.
    The task involves predicting randomly masked words and entities in a large entity-annotated corpus retrieved from Wikipedia.
    We also propose an \textit{entity-aware} self-attention mechanism that is an extension of the self-attention mechanism of the transformer, and considers the types of tokens (words or entities) when computing attention scores.
    The proposed model achieves impressive empirical performance on a wide range of entity-related tasks.
    In particular, it obtains state-of-the-art results on five well-known datasets: Open Entity (entity typing), TACRED (relation classification), CoNLL-2003 (named entity recognition), ReCoRD (cloze-style question answering), and SQuAD 1.1 (extractive question answering).
    Our source code and pretrained representations are available at \url{https://github.com/studio-ousia/luke}.
\end{abstract}

\section{Introduction}

\label{sec:introduction}

\begin{figure*}[t]
    \centering
    \includegraphics[width=\textwidth]{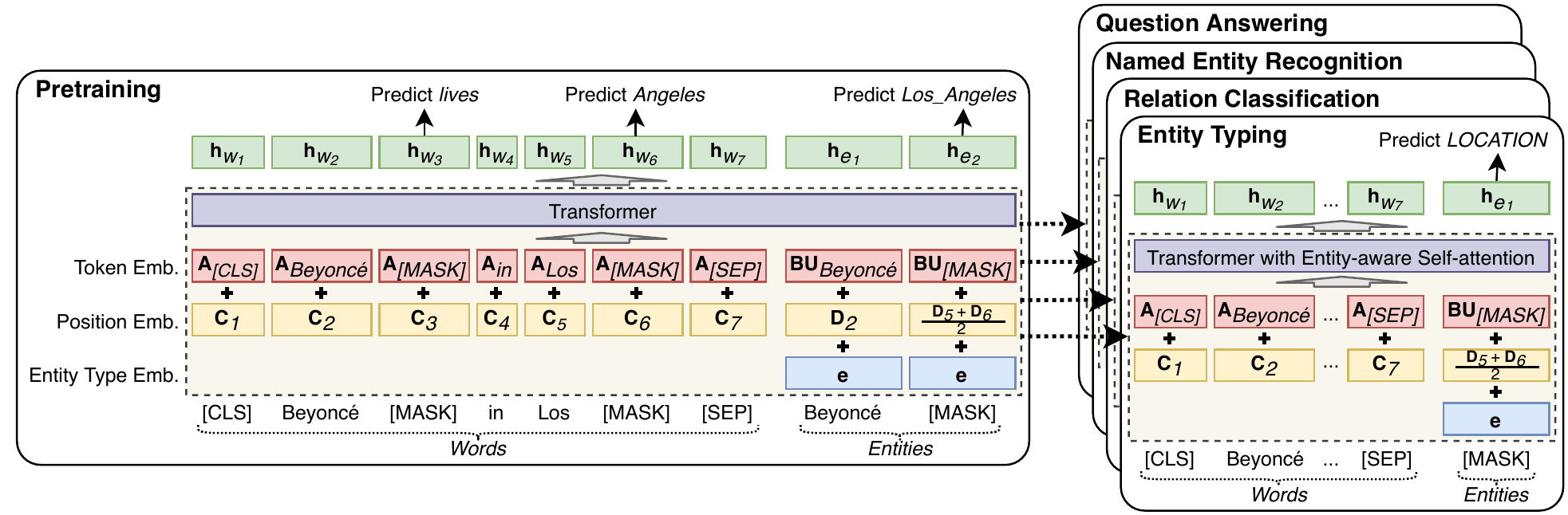}
    \caption{Architecture of LUKE using the input sentence ``\textit{Beyoncé lives in Los Angeles.}''
        LUKE outputs contextualized representation for each word and entity in the text.
        The model is trained to predict randomly masked words (e.g., \textit{lives} and \textit{Angeles} in the figure) and entities (e.g., \textit{Los\_Angeles} in the figure).
        Downstream tasks are solved using its output representations with linear classifiers.}
    \label{fig:architecture}
\end{figure*}

Many natural language tasks involve entities, e.g., relation classification, entity typing, named entity recognition (NER), and question answering (QA). Key to solving such entity-related tasks is a model to learn the effective representations of entities.
Conventional entity representations assign each entity a fixed embedding vector that stores information regarding the entity in a knowledge base (KB) \cite{Bordes2013,Trouillon2016ComplexPrediction,Yamada2016,TACL1065}.
Although these models capture the rich information in the KB, they require entity linking to represent entities in a text, and cannot represent entities that do not exist in the KB.

By contrast, contextualized word representations (CWRs) based on the transformer \cite{NIPS2017_7181}, such as BERT \cite{devlin2018bert}, and RoBERTa \cite{Liu2020RoBERTa:Approach}, provide effective general-purpose word representations trained with unsupervised pretraining tasks based on language modeling.
Many recent studies have solved entity-related tasks using the contextualized representations of entities computed based on CWRs \cite{Zhang2019,peters-knowbert,Joshi2020SpanBERT:Spans}.
However, the architecture of CWRs is not well suited to representing entities for the following two reasons:
(1) Because CWRs do not output the span-level representations of entities, they typically need to learn how to compute such representations based on a downstream dataset that is typically small.
(2) Many entity-related tasks, e.g., relation classification and QA, involve reasoning about the relationships between entities.
Although the transformer can capture the complex relationships between words by relating them to each other multiple times using the self-attention mechanism \cite{Clark2019WhatAttention,Reif2019VisualizingBERT}, it is difficult to perform such reasoning between entities because many entities are split into multiple tokens in the model.
Furthermore, the word-based pretraining task of CWRs is not suitable for learning the representations of entities because predicting a masked word given other words in the entity, e.g., predicting \textit{``Rings''} given \textit{``The Lord of the [MASK]''}, is clearly easier than predicting the entire entity.

In this paper, we propose new pretrained contextualized representations of words and entities by developing \textbf{LUKE} (\textbf{L}anguage \textbf{U}nderstanding with \textbf{K}nowledge-based \textbf{E}mbeddings).
LUKE is based on a transformer \cite{NIPS2017_7181} trained using a large amount of entity-annotated corpus obtained from Wikipedia.
An important difference between LUKE and existing CWRs is that it treats not only words, but also entities as independent tokens, and computes intermediate and output representations for all tokens using the transformer (see Figure \ref{fig:architecture}).
Since entities are treated as tokens, LUKE can directly model the relationships between entities.

LUKE is trained using a new pretraining task, a straightforward extension of BERT's masked language model (MLM) \cite{devlin2018bert}.
The task involves randomly masking entities by replacing them with \texttt{[MASK]} entities, and trains the model by predicting the originals of these masked entities.
We use RoBERTa as base pre-trained model, and conduct pretraining of the model by simultaneously optimizing the objectives of the MLM and our proposed task.
When applied to downstream tasks, the resulting model can compute representations of arbitrary entities in the text using \texttt{[MASK]} entities as inputs.
Furthermore, if entity annotation is available in the task, the model can compute entity representations based on the rich entity-centric information encoded in the corresponding entity embeddings.

Another key contribution of this paper is that it extends the transformer using our \textit{entity-aware} self-attention mechanism.
Unlike existing CWRs, our model needs to deal with two types of tokens, i.e., words and entities.
Therefore, we assume that it is beneficial to enable the mechanism to easily determine the types of tokens.
To this end, we enhance the self-attention mechanism by adopting different query mechanisms based on the attending token and the token attended to.

We validate the effectiveness of our proposed model by conducting extensive experiments on five standard entity-related tasks: entity typing,  relation classification, NER,  cloze-style QA, and extractive QA.
Our model outperforms all baseline models, including RoBERTa, in all experiments, and obtains state-of-the-art results on five tasks: entity typing on the Open Entity dataset \cite{Choi2018Ultra-FineTyping}, relation classification on the TACRED dataset \cite{Zhang2017Position-awareFilling}, NER on the CoNLL-2003 dataset \cite{TjongKimSang-DeMeulder:2003:CONLL}, cloze-style QA on the ReCoRD dataset \cite{Zhang2018ReCoRD:Comprehension}, and extractive QA on the SQuAD 1.1 dataset \cite{rajpurkar-etal-2016-squad}.
We publicize our source code and pretrained representations at \url{https://github.com/studio-ousia/luke}.

The main contributions of this paper are summarized as follows:
\begin{itemize}[leftmargin=10pt,topsep=1pt,itemsep=0pt]
    \item We propose LUKE, a new contextualized representations specifically designed to address entity-related tasks.
          LUKE is trained to predict randomly masked words and entities using a large amount of entity-annotated corpus obtained from Wikipedia.
    \item We introduce an entity-aware self-attention mechanism, an effective extension of the original mechanism of transformer. The proposed mechanism considers the type of the tokens (words or entities) when computing attention scores.
    \item LUKE achieves strong empirical performance and obtains state-of-the-art results on five popular datasets: Open Entity, TACRED, CoNLL-2003, ReCoRD, and SQuAD 1.1.
\end{itemize}

\section{Related Work}
\label{sec:related-work}

\paragraph{Static Entity Representations}

Conventional entity representations assign a fixed embedding to each entity in the KB.
They include knowledge embeddings trained on knowledge graphs \cite{Bordes2013,Yang2015EmbeddingBases,Trouillon2016ComplexPrediction}, and embeddings trained using textual contexts or descriptions of entities retrieved from a KB \cite{Yamada2016,TACL1065,cao-EtAl:2017:Long1,ganea-hofmann:2017:EMNLP2017}.
Similar to our pretraining task, \textbf{NTEE} \cite{TACL1065} and \textbf{RELIC} \cite{Ling2020LearningText} use an approach that trains entity embeddings by predicting entities given their textual contexts obtained from a KB.
The main drawbacks of this line of work, when representing entities in text, are that (1) they need to resolve entities in the text to corresponding KB entries to represent the entities, and (2) they cannot represent entities that do not exist in the KB.

\paragraph{Contextualized Word Representations}

Many recent studies have addressed entity-related tasks based on the contextualized representations of entities in text computed using the word representations of CWRs \cite{Zhang2019,BaldiniSoares2019MatchingLearning,peters-knowbert,Joshi2020SpanBERT:Spans,wang2019kepler,wang2020k}.
Representative examples of CWRs are \textbf{ELMo} \cite{peters-etal-2018-deep} and \textbf{BERT} \cite{devlin2018bert}, which are based on deep bidirectional long short-term memory (LSTM) and the transformer \cite{NIPS2017_7181}, respectively.
BERT is trained using an MLM, a pretraining task that masks random words in the text and trains the model to predict the masked words.
Most recent CWRs, such as \textbf{RoBERTa} \cite{Liu2020RoBERTa:Approach}, \textbf{XLNet} \cite{Yang2019XLNet:Understanding}, \textbf{SpanBERT} \cite{Joshi2020SpanBERT:Spans}, \textbf{ALBERT} \cite{lan2019albert}, \textbf{BART} \cite{Lewis2020BART:Comprehension}, and \textbf{T5} \cite{Raffel2020ExploringTransformer}, are based on transformer trained using a task equivalent to or similar to the MLM.
Similar to our proposed pretraining task that masks entities instead of words, several recent CWRs, e.g., SpanBERT, ALBERT, BART, and T5, have extended the MLM by randomly masking word spans instead of single words.

Furthermore, various recent studies have explored methods to enhance CWRs by injecting them with \textit{knowledge} from external sources, such as KBs.
\textbf{ERNIE} \cite{Zhang2019} and \textbf{KnowBERT} \cite{peters-knowbert} use a similar idea to enhance CWRs using static entity embeddings separately learned from a KB.
\textbf{WKLM} \cite{Xiong2020Pretrained} trains the model to detect whether an entity name in text is replaced by another entity name of the same type.
\textbf{KEPLER} \cite{wang2019kepler} conducts pretraining based on the MLM and a knowledge-embedding objective  \cite{Bordes2013}.
\textbf{K-Adapter} \cite{wang2020k} was proposed concurrently with our work, and extends CWRs using \textit{neural adapters} that inject factual and linguistic knowledge.
This line of work is related to ours because our pretraining task also enhances the model using information in the KB.

Unlike the CWRs mentioned above, LUKE uses an improved transformer architecture with an entity-aware self-attention mechanism that is designed to effectively solve entity-related tasks.
LUKE also outputs entity representations by learning how to compute them during pretraining.
It achieves superior empirical results to existing CWRs and knowledge-enhanced CWRs in all of our experiments.

\section{LUKE}

Figure \ref{fig:architecture} shows the architecture of LUKE.
The model adopts a multi-layer bidirectional transformer \cite{NIPS2017_7181}.
It treats words and entities in the document as input tokens, and computes a representation for each token.
Formally, given a sequence consisting of $m$ words $w_1, w_2, ..., w_m$ and $n$ entities $e_1, e_2, ..., e_n$, our model computes $D$-dimensional word representations $\mathbf{h}_{w_1}, \mathbf{h}_{w_2}, ..., \mathbf{h}_{w_m}$, where $\mathbf{h}_w \in \mathbb{R}^{D}$, and entity representations $\mathbf{h}_{e_1}, \mathbf{h}_{e_2}, ..., \mathbf{h}_{e_n}$, where $\mathbf{h}_e \in \mathbb{R}^{D}$.
The entities can be Wikipedia entities (e.g., \textit{Beyoncé} in Figure \ref{fig:architecture}) or special entities (e.g., \texttt{[MASK]}).

\subsection{Input Representation}
\label{sec:input_repr}

The input representation of a token (word or entity) is computed using the following three embeddings:

\begin{itemize}[leftmargin=10pt,topsep=3pt,itemsep=3pt]
    \item \textbf{Token embedding} represents the corresponding token.
          We denote the word token embedding by $\mathbf{A} \in \mathbb{R}^{V_w\times D}$, where $V_w$ is the number of words in our vocabulary.
          For computational efficiency, we represent the entity token embedding by decomposing it into two small matrices, $\mathbf{B} \in \mathbb{R}^{V_{e}\times H}$ and $\mathbf{U} \in \mathbb{R}^{H\times D}$, where $V_e$ is the number of entities in our vocabulary.
          Hence, the full matrix of the entity token embedding can be computed as $\mathbf{B}\mathbf{U}$.

    \item \textbf{Position embedding} represents the position of the token in a word sequence.
          A word and an entity appearing at the $i$-th position in the sequence are represented as $\mathbf{C}_i \in \mathbb{R}^D$ and $\mathbf{D}_i \in \mathbb{R}^D$, respectively.
          If an entity name contains multiple words, its position embedding is computed by averaging the embeddings of the corresponding positions, as shown in Figure \ref{fig:architecture}.

    \item \textbf{Entity type embedding} represents that the token is an entity.
        The embedding is a single vector denoted by $\mathbf{e} \in \mathbb{R}^D$.
\end{itemize}

The input representation of a word and that of an entity are computed by summing the token and position embeddings, and the token, position, and entity type embeddings, respectively.
Following past work \cite{devlin2018bert,Liu2020RoBERTa:Approach}, we insert special tokens \texttt{[CLS]} and \texttt{[SEP]} into the word sequence as the first and last words, respectively.

\subsection{Entity-aware Self-attention}
\label{sec:entity_aware_attn}

The self-attention mechanism is the foundation of the transformer \cite{NIPS2017_7181}, and relates tokens each other based on the attention score between each pair of tokens.
Given a sequence of input vectors $\mathbf{x}_1, \mathbf{x}_2, ..., \mathbf{x}_k$, where $\mathbf{x}_i \in \mathbb{R}^D$, each of the output vectors $\mathbf{y}_1, \mathbf{y}_2, ..., \mathbf{y}_k$, where $\mathbf{y}_i \in \mathbb{R}^L$, is computed based on the weighted sum of the transformed input vectors.
Here, each input and output vector corresponds to a token (a word or an entity) in our model; therefore, $k = m + n$.
The $i$-th output vector $\mathbf{y}_i$ is computed as:
\begin{align*}
    \mathbf{y}_i & = \sum^k_{j=1}\alpha_{ij}\mathbf{V}\mathbf{x}_j                      \\
    e_{ij}       & = \frac{\mathbf{K}\mathbf{x}_j^\top\mathbf{Q}\mathbf{x}_i}{\sqrt{L}} \\
    \alpha_{ij}  & = \mathrm{softmax}(e_{ij})
\end{align*}
where $\mathbf{Q} \in \mathbb{R}^{L\times D}$, $\mathbf{K} \in \mathbb{R}^{L\times D}$, and $\mathbf{V} \in \mathbb{R}^{L\times D}$ denote the query, key, and value matrices, respectively.

Because LUKE handles two types of tokens (i.e., words and entities), we assume that it is beneficial to use the information of target token types when computing the attention scores ($e_{ij}$).
With this in mind, we enhance the mechanism by introducing an \textit{entity-aware} query mechanism that uses a different query matrix for each possible pair of token types of $\mathbf{x}_i$ and $\mathbf{x}_j$.
Formally, the attention score $e_{ij}$ is computed as follows:
\begin{equation*}
    \resizebox{\linewidth}{!}{$\displaystyle
            e_{ij}=
            \begin{cases}
                \mathbf{K}\mathbf{x}_j^\top\mathbf{Q}\mathbf{x}_i,       & \text{\small{if both }} \mathbf{x}_i \text{\small{ and }} \mathbf{x}_j \text{\small{ are words}}    \\
                \mathbf{K}\mathbf{x}_j^\top\mathbf{Q}_{w2e}\mathbf{x}_i, & \text{\small{if }} \mathbf{x}_i \text{\small{ is word and }} \mathbf{x}_j \text{\small{ is entity}} \\
                \mathbf{K}\mathbf{x}_j^\top\mathbf{Q}_{e2w}\mathbf{x}_i, & \text{\small{if }} \mathbf{x}_i \text{\small{ is entity and }} \mathbf{x}_j \text{\small{ is word}} \\
                \mathbf{K}\mathbf{x}_j^\top\mathbf{Q}_{e2e}\mathbf{x}_i, & \text{\small{if both }} \mathbf{x}_i \text{\small{ and }} \mathbf{x}_j \text{\small{ are entities}} \\
            \end{cases}
        $}
\end{equation*}
where $\mathbf{Q}_{w2e}$, $\mathbf{Q}_{e2w}$, $\mathbf{Q}_{e2e} \in \mathbb{R}^{L\times D}$ are query matrices.
Note that the computational costs of the original mechanism and our proposed mechanism are identical except the additional cost of computing gradients and updating the parameters of the additional query matrices at the training time.

\subsection{Pretraining Task}
\label{subsec:pretraining}

To pretrain LUKE, we use the conventional MLM and a new pretraining task that is an extension of the MLM to learn entity representations.
In particular, we treat hyperlinks in Wikipedia as entity annotations, and train the model using a large entity-annotated corpus retrieved from Wikipedia.
We randomly mask a certain percentage of the entities by replacing them with special \texttt{[MASK]} entities\footnote{Note that LUKE uses two different \texttt{[MASK]} tokens: the \texttt{[MASK]} word for MLM and the \texttt{[MASK]} entity for our proposed pretraining task.} and then train the model to predict the masked entities.
Formally, the original entity corresponding to a masked entity is predicted by applying the softmax function over all entities in our vocabulary:
\begin{align*}
     & \mathbf{\hat{y}} = \mathrm{softmax}(\mathbf{B}\mathbf{T}\mathbf{m} + \mathbf{b}_{o})              \\
     & \mathbf{m} = \mathrm{layer\_norm}\big(\mathrm{gelu}(\mathbf{W}_h\mathbf{h}_e + \mathbf{b}_h)\big)
\end{align*}
where $\mathbf{h}_e$ is the representation corresponding to the masked entity, $\mathbf{T} \in \mathbb{R}^{H\times D}$ and $\mathbf{W}_h \in \mathbb{R}^{D\times D}$ are weight matrices, $\mathbf{b}_{o} \in \mathbb{R}^{V_e}$ and $\mathbf{b}_h \in \mathbb{R}^D$ are bias vectors, $\text{gelu}(\cdot)$ is the gelu activation function \cite{hendrycks2016gaussian}, and $\text{layer\_norm}(\cdot)$ is the layer normalization function \cite{lei2016layer}.
Our final loss function is the sum of MLM loss and cross-entropy loss on predicting the masked entities, where the latter is computed identically to the former.

\subsection{Modeling Details}
\label{subsec:modelling-details}

Our model configuration follows RoBERTa$_{\text{LARGE}}$ \cite{Liu2020RoBERTa:Approach}, pretrained CWRs based on a bidirectional transformer and a variant of BERT \cite{devlin2018bert}.
In particular, our model is based on the bidirectional transformer with $D = 1024$ hidden dimensions, 24 hidden layers, $L = 64$ attention head dimensions, and 16 self-attention heads.
The number of dimensions of the entity token embedding is set to $H = 256$.
The total number of parameters is approximately 483 M, consisting of 355 M in RoBERTa and 128 M in our entity embeddings.
The input text is tokenized into words using RoBERTa's tokenizer with the vocabulary consisting of $V_w = 50K$ words.
For computational efficiency, our entity vocabulary does not include all entities but only the $V_e = 500K$ entities most frequently appearing in our entity annotations.
The entity vocabulary also includes two special entities, i.e., \texttt{[MASK]} and \texttt{[UNK]}.

The model is trained via iterations over Wikipedia pages in a random order for 200K steps.
To reduce training time, we initialize the parameters that LUKE have in common with RoBERTa (parameters in the transformer and the embeddings for words) using RoBERTa.
Following past work \cite{devlin2018bert,Liu2020RoBERTa:Approach}, we mask 15\% of all words and entities at random.
If an entity does not exist in the vocabulary, we replace it with the \texttt{[UNK]} entity.
We perform pretraining using the original self-attention mechanism rather than our entity-aware self-attention mechanism because we want an ablation study of our mechanism but can not afford to run pretraining twice.
Query matrices of our self-attention mechanism ($\mathbf{Q}_{w2e}$, $\mathbf{Q}_{e2w}$, and $\mathbf{Q}_{e2e}$) are learned using downstream datasets.
Further details of our pretraining are described in Appendix \ref{sec:detail-pretraining}.

\section{Experiments}

We conduct extensive experiments using five entity-related tasks: entity typing, relation classification, NER, cloze-style QA, and extractive QA.
We use similar model architectures for all tasks based on a simple linear classifier on top of the representations of words, entities, or both.
Unless otherwise specified, we create the input word sequence by inserting tokens of \texttt{[CLS]} and \texttt{[SEP]} into the original word sequence as the first and the last tokens, respectively.
The input entity sequence is built using \texttt{[MASK]} entities, special entities introduced for the task, or Wikipedia entities.
The token embedding of a task-specific special entity is initialized using that of the \texttt{[MASK]} entity, and the query matrices of our entity-aware self-attention mechanism ($\mathbf{Q}_{w2e}$, $\mathbf{Q}_{e2w}$, and $\mathbf{Q}_{e2e}$) are initialized using the original query matrix $\mathbf{Q}$.

Because we use RoBERTa as the base model in our pretraining, we use it as our primary baseline for all tasks.
We omit a description of the baseline models in each section if they are described in Section \ref{sec:related-work}.
Further details of our experiments are available in Appendix \ref{sec:detail-fine-tuning}.

\subsection{Entity Typing}

We first conduct experiments on entity typing, which is the task of predicting the types of an entity in the given sentence.
Following \newcite{Zhang2019}, we use the Open Entity dataset \cite{Choi2018Ultra-FineTyping}, and consider only nine general entity types.
Following \newcite{wang2020k}, we report loose micro-precision, recall, and F1, and employ the micro-F1 as the primary metric.

\paragraph{Model}

We represent the target entity using the \texttt{[MASK]} entity, and enter words and the entity in each sentence into the model.
We then classify the entity using a linear classifier based on the corresponding entity representation.
We treat the task as multi-label classification, and train the model using binary cross-entropy loss averaged over all entity types.

\paragraph{Baselines}

\textbf{UFET} \cite{Choi2018Ultra-FineTyping} is a conventional model that computes context representations using the bidirectional LSTM.
We also use BERT, RoBERTa, ERNIE, KnowBERT, KEPLER, and K-Adapter as baselines.

\paragraph{Results}

\begin{table}[t]
    \centering
    \setlength{\tabcolsep}{4pt}
    \scalebox{0.75}{
        \begin{tabular}{lccc}
            \hline
            Name                            & Prec.         & Rec.          & F1            \\
            \hline
            UFET \cite{Zhang2019}           & 77.4          & 60.6          & 68.0          \\
            BERT \cite{Zhang2019}           & 76.4          & 71.0          & 73.6          \\
            ERNIE \cite{Zhang2019}          & 78.4          & 72.9          & 75.6          \\
            KEPLER \cite{wang2019kepler}    & 77.2          & 74.2          & 75.7          \\
            KnowBERT \cite{peters-knowbert} & 78.6          & 73.7          & 76.1          \\
            K-Adapter \cite{wang2020k}      & 79.3          & 75.8          & 77.5          \\
            \hline
            RoBERTa \cite{wang2020k}        & 77.6          & 75.0          & 76.2          \\
            LUKE                            & \textbf{79.9} & \textbf{76.6} & \textbf{78.2} \\
            \hline
        \end{tabular}
    }
    \caption{Results of entity typing on the Open Entity dataset.}
    \label{tb:open-entity-results}
\end{table}

Table \ref{tb:open-entity-results} shows the experimental results.
LUKE significantly outperforms our primary baseline, RoBERTa, by 2.0 F1 points, and the previous best published model, KnowBERT, by 2.1 F1 points.
Furthermore, LUKE achieves a new state of the art by outperforming K-Adapter by 0.7 F1 points.

\subsection{Relation Classification}

Relation classification determines the correct relation between \textit{head} and \textit{tail} entities in a sentence.
We conduct experiments using TACRED dataset \cite{Zhang2017Position-awareFilling}, a large-scale relation classification dataset containing 106,264 sentences with 42 relation types.
Following \newcite{wang2020k}, we report the micro-precision, recall, and F1, and use the micro-F1 as the primary metric.

\paragraph{Model}

We introduce two special entities, \texttt{[HEAD]} and \texttt{[TAIL]}, to represent the head and the tail entities, respectively, and input words and these two entities in each sentence to the model.
We then solve the task using a linear classifier based on a concatenated representation of the head and tail entities.
The model is trained using cross-entropy loss.

\paragraph{Baselines}

\textbf{C-GCN} \cite{Zhang2018GraphExtraction} uses graph convolutional networks over dependency tree structures to solve the task.
\textbf{MTB} \cite{BaldiniSoares2019MatchingLearning} learns relation representations based on BERT through the \textit{matching-the-blanks} task using a large amount of entity-annotated text.
We also compare LUKE with BERT, RoBERTa, SpanBERT, ERNIE, KnowBERT, KEPLER, and K-Adapter.

\paragraph{Results}

\begin{table}[t]
    \centering
    \setlength{\tabcolsep}{4pt}
    \scalebox{0.75}{
        \begin{tabular}{lccc}
            \hline
            Name                                         & Prec.         & Rec.          & F1            \\
            \hline
            BERT \cite{Zhang2019}                        & 67.2          & 64.8          & 66.0          \\
            C-GCN \cite{Zhang2018GraphExtraction}        & 69.9          & 63.3          & 66.4          \\
            ERNIE \cite{Zhang2019}                       & 70.0          & 66.1          & 68.0          \\
            SpanBERT \cite{Joshi2020SpanBERT:Spans}      & 70.8          & 70.9          & 70.8          \\
            MTB \cite{BaldiniSoares2019MatchingLearning} & -             & -             & 71.5          \\
            KnowBERT \cite{peters-knowbert}              & \textbf{71.6} & 71.4          & 71.5          \\
            KEPLER \cite{wang2019kepler}                 & 70.4          & 73.0          & 71.7          \\
            K-Adapter \cite{wang2020k}                   & 68.9          & \textbf{75.4} & 72.0          \\
            \hline
            RoBERTa \cite{wang2020k}                     & 70.2          & 72.4          & 71.3          \\
            LUKE                                         & 70.4          & 75.1          & \textbf{72.7} \\
            \hline
        \end{tabular}
    }
    \caption{Results of relation classification on the TACRED dataset.}
    \label{tb:tacred-results}
\end{table}

The experimental results are presented in Table \ref{tb:tacred-results}.
LUKE clearly outperforms our primary baseline, RoBERTa, by 1.4 F1 points, and the previous best published models, namely MTB and KnowBERT, by 1.2 F1 points.
Furthermore, it achieves a new state of the art by outperforming K-Adapter by 0.7 F1 points.

\subsection{Named Entity Recognition}
\label{subsec:ner}

We conduct experiments on the NER task using the standard CoNLL-2003 dataset \cite{TjongKimSang-DeMeulder:2003:CONLL}.
Following past work, we report the span-level F1.

\paragraph{Model}

Following \newcite{Sohrab2018DeepRecognition}, we solve the task by enumerating all possible spans (or n-grams) in each sentence as entity name candidates, and classifying them into the target entity types or \textit{non-entity} type, which indicates that the span is not an entity.
For each sentence in the dataset, we enter words and the \texttt{[MASK]} entities corresponding to all possible spans.
The representation of each span is computed by concatenating the word representations of the first and last words in the span, and the entity representation corresponding to the span.
We classify each span using a linear classifier with its representation, and train the model using cross-entropy loss.
We exclude spans longer than 16 words for computational efficiency.
During the inference, we first exclude all spans classified into the \textit{non-entity} type.
To avoid selecting overlapping spans, we greedily select a span from the remaining spans based on the logit of its predicted entity type in descending order if the span does not overlap with those already selected.
Following \newcite{devlin2018bert}, we include the maximal document context in the target document.

\paragraph{Baselines}

\textbf{LSTM-CRF} \cite{Lample2016NeuralRecognition} is a model based on the bidirectional LSTM with conditional random fields (CRF).
\textbf{\newcite{Akbik2018ContextualLabeling}} address the task using the bidirectional LSTM with CRF enhanced with character-level contextualized representations.
Similarly, \textbf{\newcite{Baevski2019Cloze-drivenNetworks}} use the bidirectional LSTM with CRF enhanced with CWRs based on a bidirectional transformer.
We also use ELMo, BERT, and RoBERTa as baselines.
To conduct a fair comparison with RoBERTa, we report its performance using the model described above with the span representation computed by concatenating the representations of the first and last words of the span.

\paragraph{Results}

\begin{table}[t]
    \centering
    \setlength{\tabcolsep}{4pt}
    \scalebox{0.75}{
        \begin{tabular}{lc}
            \hline
            Name                                        & F1            \\
            \hline
            LSTM-CRF \cite{Lample2016NeuralRecognition} & 91.0          \\
            ELMo \cite{peters-etal-2018-deep}           & 92.2          \\
            BERT \cite{devlin2018bert}                  & 92.8          \\
            \newcite{Akbik2018ContextualLabeling}       & 93.1          \\
            \newcite{Baevski2019Cloze-drivenNetworks}   & 93.5          \\
            \hline
            RoBERTa                                     & 92.4          \\
            LUKE                                        & \textbf{94.3} \\
            \hline
        \end{tabular}
    }
    \caption{Results of named entity recognition on the CoNLL-2003 dataset.}
    \label{tb:ner-results}
\end{table}

The experimental results are shown in Table \ref{tb:ner-results}.
LUKE outperforms RoBERTa by 1.9 F1 points.
Furthermore, it achieves a new state of the art on this competitive dataset by outperforming the previous state of the art reported in \newcite{Baevski2019Cloze-drivenNetworks} by 0.8 F1 points.

\subsection{Cloze-style Question Answering}

We evaluate our model on the ReCoRD dataset \cite{Zhang2018ReCoRD:Comprehension}, a cloze-style QA dataset consisting of over 120K examples.
An interesting characteristic of this dataset is that most of its questions cannot be solved without external knowledge.
The following is an example question and its answer in the dataset:
\begin{quoting}[vskip=0pt,leftmargin=0.5em,rightmargin=0.5em]
    \textbf{Question:} According to claims in the suit, ``Parts of 'Stairway to Heaven,' instantly recognizable to the music fans across the world, sound almost identical to significant portions of ‘\textbf{X}.’''\\
    \textbf{Answer:} Taurus
\end{quoting}
Given a question and a passage, the task is to find the entity mentioned in the passage that fits the missing entity (denoted by \textbf{X} in the question above).
In this dataset, annotations of entity spans (start and end positions) in a passage are provided, and the answer is contained in the provided entity spans one or multiple times.
Following past work, we evaluate the models using exact match (EM) and token-level F1 on the development and test sets.

\paragraph{Model}

We solve this task by assigning a relevance score to each entity in the passage and selecting the entity with the highest score as the answer.
Following \newcite{Liu2020RoBERTa:Approach}, given a question $q_1, q_2, ..., q_j$, and a passage $p_1, p_2, ..., p_l$, the input word sequence is constructed as: \texttt{[CLS]}$q_1, q_2, ..., q_j$\texttt{[SEP]} \texttt{[SEP]}$p_1, p_2, ..., p_l$\texttt{[SEP]}.
Further, we input \texttt{[MASK]} entities corresponding to the missing entity and all entities in the passage.
We compute the relevance score of each entity in the passage using a linear classifier with the concatenated representation of the missing entity and the corresponding entity.
We train the model using binary cross-entropy loss averaged over all entities in the passage, and select the entity with the highest score (logit) as the answer.

\paragraph{Baselines}

\textbf{DocQA+ELMo} \cite{Clark2018SimpleComprehension} is a model based on ELMo, bidirectional attention flow \cite{Seo2017BidirectionalComprehension}, and self-attention mechanism.
\textbf{XLNet+Verifier} \cite{Li2019PinganTasks} is a model based on XLNet with rule-based answer verification, and is the winner of a recent competition based on this dataset \cite{Ostermann2019CommonsenseReport}.
We also use BERT and RoBERTa as baselines.

\paragraph{Results}

\begin{table}[t]
    \centering
    \setlength{\tabcolsep}{3pt}
    \scalebox{0.75}{
        \begin{tabular}{lcccc}
            \hline
            Name                                              & \makecell{Dev\\EM} & \makecell{Dev\\F1} & \makecell{Test\\EM} & \makecell{Test\\F1}\\
            \hline
            DocQA+ELMo \cite{Zhang2018ReCoRD:Comprehension}   & 44.1          & 45.4          & 45.4          & 46.7          \\
            BERT \cite{Wang2019SuperGLUE:Systems}             & -             & -             & 71.3          & 72.0          \\
            XLNet+Verifier \cite{Li2019PinganTasks}           & 80.6          & 82.1          & 81.5          & 82.7          \\
            \hline
            RoBERTa \cite{Liu2020RoBERTa:Approach}            & 89.0          & 89.5          & -             & -             \\
            RoBERTa (ensemble) \cite{Liu2020RoBERTa:Approach} & -             & -             & 90.0          & 90.6          \\
            LUKE                                              & \textbf{90.8} & \textbf{91.4} & \textbf{90.6} & \textbf{91.2} \\
            \hline
        \end{tabular}
    }
    \caption{Results of cloze-style question answering on the ReCoRD dataset. All models except RoBERTa (ensemble) are based on a single model.}
    \label{tb:record-results}
\end{table}

The results are presented in Table \ref{tb:record-results}.
LUKE significantly outperforms RoBERTa, the best baseline, on the development set by 1.8 EM points and 1.9 F1 points.
Furthermore, it achieves superior results to RoBERTa (ensemble) on the test set without ensembling the models.

\subsection{Extractive Question Answering}
\label{subsec:squad}

Finally, we conduct experiments using the well-known Stanford Question Answering Dataset (SQuAD) 1.1 consisting of 100K question/answer pairs \cite{rajpurkar-etal-2016-squad}.
Given a question and a Wikipedia passage containing the answer, the task is to predict the answer span in the passage.
Following past work, we report the EM and token-level F1 on the development and test sets.

\paragraph{Model}

We construct the word sequence from the question and the passage in the same way as in the previous experiment.
Unlike in the other experiments, we input Wikipedia entities into the model based on entity annotations automatically generated on the question and the passage using a mapping from entity names (e.g., ``U.S.'') to their referent entities (e.g., \textit{United States}).
The mapping is automatically created using the entity hyperlinks in Wikipedia as described in detail in Appendix \ref{sec:squad-annotations}.
We solve this task using the same model architecture as that of BERT and RoBERTa.
In particular, we use two linear classifiers independently on top of the word representations to predict the span boundary of the answer (i.e., the start and end positions), and train the model using cross-entropy loss.

\paragraph{Baselines}

We compare our models with the results of recent CWRs, including BERT, RoBERTa, SpanBERT, XLNet, and ALBERT.
Because the results for RoBERTa and ALBERT are reported only on the development set, we conduct a comparison with these models using this set.
To conduct a fair compassion with RoBERTa, we use the same model architecture and hyper-parameters as those of RoBERTa \cite{Liu2020RoBERTa:Approach}.

\paragraph{Results}

\begin{table}[t]
    \centering
    \setlength{\tabcolsep}{3pt}
    \scalebox{0.75}{
        \begin{tabular}{lcccc}
            \hline
            Name                                     & \makecell{Dev\\EM} & \makecell{Dev\\F1} & \makecell{Test\\EM} & \makecell{Test\\F1}\\
            \hline
            BERT \cite{devlin2018bert}               & 84.2          & 91.1          & 85.1          & 91.8          \\
            SpanBERT \cite{Joshi2020SpanBERT:Spans}  & -             & -             & 88.8          & 94.6          \\
            XLNet \cite{Yang2019XLNet:Understanding} & 89.0          & 94.5          & 89.9          & 95.1          \\
            ALBERT \cite{lan2019albert}              & 89.3          & 94.8          & -             & -             \\
            \hline
            RoBERTa \cite{Liu2020RoBERTa:Approach}   & 88.9          & 94.6          & -             & -             \\
            LUKE                                     & \textbf{89.8} & \textbf{95.0} & \textbf{90.2} & \textbf{95.4} \\
            \hline
        \end{tabular}
    }
    \caption{Results of extractive question answering on the SQuAD 1.1 dataset.}
    \label{tb:squad-results}
\end{table}

\begin{table}[t]
    \centering
    \setlength{\tabcolsep}{3pt}
    \scalebox{0.75}{
        \begin{tabular}{lccc}
            \hline
            Name                   & \makecell{CoNLL-2003\\(Test F1)}&\makecell{SQuAD\\(Dev EM)} & \makecell{SQuAD\\(Dev F1)}\\
            \hline
            LUKE w/o entity inputs & 92.9                 & 89.2          & 94.8          \\
            LUKE                   & \textbf{94.3}        & \textbf{89.8} & \textbf{95.0} \\
            \hline
        \end{tabular}
    }
    \caption{Ablation study of our entity representations.}
    \label{tb:luke-ablation-study}
\end{table}

\begin{table*}[t]
    \centering
    \setlength{\tabcolsep}{3pt}
    \scalebox{0.75}{
        \begin{tabular}{lccccccc}
            \hline
            Name                   & \makecell{Open Entity\\(Test F1)} & \makecell{TACRED\\(Test F1)} & \makecell{CoNLL-2003\\(Test F1)} & \makecell{ReCoRD\\(Dev EM)} & \makecell{ReCoRD\\(Dev F1)} & \makecell{SQuAD\\(Dev EM)} & \makecell{SQuAD\\(Dev F1)}\\
            \hline
            Original Attention     & 77.9                  & 72.2          & 94.1          & 90.1          & 90.7          & 89.2          & 94.7          \\
            Entity-aware Attention & \textbf{78.2}         & \textbf{72.7} & \textbf{94.3} & \textbf{90.8} & \textbf{91.4} & \textbf{89.8} & \textbf{95.0} \\
            \hline
        \end{tabular}
    }
    \caption{Ablation study of our entity-aware self-attention mechanism.}
    \label{tb:entity-aware-attention-study}
\end{table*}

\begin{table}[t]
    \centering
    \setlength{\tabcolsep}{3pt}
    \scalebox{0.75}{
        \begin{tabular}{lccc}
            \hline
            Name                      & \makecell{CoNLL-2003\\(Test F1)}&\makecell{SQuAD\\(Dev EM)} & \makecell{SQuAD\\(Dev F1)}\\
            \hline
            RoBERTa w/ extra training & 92.5                 & 89.1          & 94.7          \\
            RoBERTa                   & 92.4                 & 88.9          & 94.6          \\
            LUKE                      & \textbf{94.3}        & \textbf{89.8} & \textbf{95.0} \\
            \hline
        \end{tabular}
    }
    \caption{Results of RoBERTa additionally trained using our Wikipedia corpus.}
    \label{tb:roberta-wiki-study}
\end{table}

The experimental results are presented in Table \ref{tb:squad-results}.
LUKE outperforms our primary baseline, RoBERTa, by 0.9 EM points and 0.4 F1 points on the development set.
Furthermore, it achieves a new state of the art on this competitive dataset by outperforming XLNet by 0.3 points both in terms of EM and F1.
Note that XLNet uses a more sophisticated model involving beam search than the other models considered here.

\section{Analysis}
\label{sec:analysis}

In this section, we provide a detailed analysis of LUKE by reporting three additional experiments.

\subsection{Effects of Entity Representations}

To investigate how our entity representations influence performance on downstream tasks, we perform an ablation experiment by addressing NER on the CoNLL-2003 dataset and extractive QA on the SQuAD dataset without inputting any entities.
In this setting, LUKE uses only the word sequence to compute the representation for each word.
We address the tasks using the same model architectures as those for RoBERTa described in the corresponding sections.
As shown in Table \ref{tb:luke-ablation-study}, this setting clearly degrades performance, i.e., 1.4 F1 points on the CoNLL-2003 dataset and 0.6 EM points on the SQuAD dataset, demonstrating the effectiveness of our entity representations on these two tasks.

\subsection{Effects of Entity-aware Self-attention}

We conduct an ablation study of our entity-aware self-attention mechanism by comparing the performance of LUKE using our mechanism with that using the original mechanism of the transformer.
As shown in Table \ref{tb:entity-aware-attention-study}, our entity-aware self-attention mechanism consistently outperforms the original mechanism across all tasks.
Furthermore, we observe significant improvements on two kinds of tasks, relation classification (TACRED) and QA (ReCoRD and SQuAD).
Because these tasks involve reasoning based on relationships between entities, we consider that our mechanism enables the model (i.e., attention heads) to easily focus on capturing the relationships between entities.

\subsection{Effects of Extra Pretraining}

As mentioned in Section \ref{subsec:modelling-details}, LUKE is based on RoBERTa with pretraining for 200K steps using our Wikipedia corpus.
Because past studies \cite{Liu2020RoBERTa:Approach,lan2019albert} suggest that simply increasing the number of training steps of CWRs tends to improve performance on downstream tasks, the superior experimental results of LUKE compared with those of RoBERTa may be obtained because of its greater number of pretraining steps.
To investigate this, we train another model based on RoBERTa with extra pretraining based on the MLM using the Wikipedia corpus for 200K training steps.
The detailed configuration used in the pretraining is available in Appendix \ref{sec:detail-pretraining}.

We evaluate the performance of this model on the CoNLL-2003 and SQuAD datasets using the same model architectures as those for RoBERTa described in the corresponding sections.
As shown in Table \ref{tb:roberta-wiki-study}, the model achieves similar performance to the original RoBERTa on both datasets, which indicates that the superior performance of LUKE is not owing to its longer pretraining.

\section{Conclusions}

In this paper, we propose LUKE, new pretrained contextualized representations of words and entities based on the transformer.
LUKE outputs the contextualized representations of words and entities using an improved transformer architecture with using a novel entity-aware self-attention mechanism.
The experimental results prove its effectiveness on various entity-related tasks.
Future work involves applying LUKE to domain-specific tasks, such as those in biomedical and legal domains.

\bibliography{references}
\bibliographystyle{acl_natbib}

\appendix

\section*{Appendix for ``LUKE: Deep Contextualized Entity Representations with Entity-aware Self-attention''}

\section{Details of Pretraining}
\label{sec:detail-pretraining}

As input corpus for pretraining, we use the December 2018 version of Wikipedia, comprising approximately 3.5 billion words and 11 million entity annotations.
We generate input sequences by splitting the content of each page into sequences comprising $\leq 512$ words and their entity annotations (i.e., hyperlinks).
We optimize the model using AdamW with learning rate warmup and linear decay of the learning rate.
To stabilize training, we update only those parameters that are randomly initialized (i.e., fix the parameters that are initialized using RoBERTa) in the first 100K steps, and update all parameters in the remaining 100K steps.
We run the pretraining on NVIDIA's PyTorch Docker container 19.02 hosted on a server with two Intel Xeon Platinum 8168 CPUs and 16 NVIDIA Tesla V100 GPUs.
The training takes approximately 30 days.
The detailed hyper-parameters are shown in Table  \ref{tb:pretraining-config}.

Table \ref{tb:roberta-pretraining-config} shows the hyper-parameters used for the extra pretraining of RoBERTa on our Wikipedia corpus described in Section \ref{sec:analysis}.
As shown in the Table, we use the same hyper-parameters as the ones used to train LUKE.
We train the model for 200K steps and update all parameters throughout the training.

\begin{table}[tb]
    \centering
    \scalebox{0.75}{
        \begin{tabular}{l|c}
            \hline
            Name                                  & Value  \\
            \hline
            Maximum word length                   & 512    \\
            Batch size                            & 2048   \\
            Peak learning rate                    & 1e-5   \\
            Peak learning rate (first 100K steps) & 5e-4   \\
            Learning rate decay                   & linear \\
            Warmup steps                          & 2500   \\
            Mask probability for words            & 15\%   \\
            Mask probability for entities         & 15\%   \\
            Dropout                               & 0.1    \\
            Weight decay                          & 0.01   \\
            Gradient clipping                     & none   \\
            Adam $\beta_1$                        & 0.9    \\
            Adam $\beta_2$                        & 0.999  \\
            Adam $\epsilon$                       & 1e-6   \\
            \hline
        \end{tabular}
    }
    \caption{Hyper-parameters used to pretrain LUKE.}
    \label{tb:pretraining-config}
\end{table}

\begin{table}[tb]
    \centering
    \scalebox{0.75}{
        \begin{tabular}{l|c}
            \hline
            Name                       & Value  \\
            \hline
            Maximum word length        & 512    \\
            Batch size                 & 2048   \\
            Peak learning rate         & 1e-5   \\
            Learning rate decay        & linear \\
            Warmup steps               & 2500   \\
            Mask probability for words & 15\%   \\
            Dropout                    & 0.1    \\
            Weight decay               & 0.01   \\
            Gradient clipping          & none   \\
            Adam $\beta_1$             & 0.9    \\
            Adam $\beta_2$             & 0.999  \\
            Adam $\epsilon$            & 1e-6   \\
            \hline
        \end{tabular}
    }
    \caption{Hyper-parameters used for the extra pretraining of RoBERTa on our Wikipedia corpus.}
    \label{tb:roberta-pretraining-config}
\end{table}

\section{Details of Experiments}
\label{sec:detail-fine-tuning}

\begin{table*}[tb]
    \centering
    \scalebox{0.75}{
        \begin{tabular}{l|ccccc}
            \hline
            Name            & Open Entity & TACRED  & CoNLL-2003 & ReCoRD          & SQuAD           \\
            \hline
            Learning rate   & 1e-5        & 1e-5    & 1e-5       & 1e-5            & 15e-6           \\
            Batch size      & 4           & 32      & 8          & 32              & 48              \\
            Training epochs & 3           & 5       & 5          & 2               & 2               \\
            Training time   & 10min       & 190min  & 203min     & 92min           & 42min           \\
            Number of GPUs  & 1           & 1       & 1          & 8               & 8               \\
            Dev score       & 78.5 F1     & 72.0 F1 & 97.1 F1    & 90.8 EM/91.4 F1 & 89.8 EM/95.0 F1 \\
            \hline
        \end{tabular}
    }
    \caption{Hyper-parameters and other details of our experiments.}
    \label{tb:fine-tuning-hyperparam-config}
\end{table*}

\begin{table}[tb]
    \centering
    \scalebox{0.75}{
        \begin{tabular}{l|c}
            \hline
            Name                & Value  \\
            \hline
            Maximum word length & 512    \\
            Learning rate decay & linear \\
            Warmup ratio        & 0.06   \\
            Dropout             & 0.1    \\
            Weight decay        & 0.01   \\
            Gradient clipping   & none   \\
            Adam $\beta_1$      & 0.9    \\
            Adam $\beta_2$      & 0.98   \\
            Adam $\epsilon$     & 1e-6   \\
            \hline
        \end{tabular}
    }
    \caption{Common hyper-parameters used in our experiments.}
    \label{tb:fine-tuning-common-config}
\end{table}

We conduct the experiments using NVIDIA's PyTorch Docker container 19.02 hosted on a server with two Intel Xeon E5-2698 v4 CPUs and eight V100 GPUs.
For each dataset, excluding SQuAD, we conduct hyper-parameter tuning using grid search based on the performance on the development set.
We evaluate performance using EM on the ReCoRD dataset, and F1 on the other datasets.
Because our computational resources are limited, we use the following constrained search space:
\begin{itemize}[leftmargin=10pt,topsep=1pt,itemsep=0pt]
    \item learning rate: 1e-5, 2e-5, 3e-5
    \item batch size: 4, 8, 16, 32, 64
    \item number of training epochs: 2, 3, 5
\end{itemize}
We do not tune the hyper-parameters of the SQuAD dataset, and use the ones described in \newcite{Liu2020RoBERTa:Approach}.
The hyper-parameters and other details, including the training time, number of GPUs used, and the best score on the development set, are shown in Table \ref{tb:fine-tuning-hyperparam-config}.
For the other hyper-parameters, we simply follow \newcite{Liu2020RoBERTa:Approach} (see Table \ref{tb:fine-tuning-common-config}).
We optimize the model using AdamW with learning rate warmup and linear decay of the learning rate.
We also use early stopping based on performance on the development set.
The details of the datasets used in our experiments are provided below.

\subsection{Open Entity}

The Open Entity dataset used in \newcite{Zhang2019} consists of training, development, and test sets, where each set contains 1,998 examples with labels of nine general entity types.
The dataset is downloaded from the website for \newcite{Zhang2019}.\footnote{\url{https://github.com/thunlp/ERNIE}}
We compute the reported results using our code based on that of \newcite{Zhang2019}.

\subsection{TACRED}

The TACRED dataset contains 68,124 training examples, 22,631 development examples, and 15,509 test examples with labels of their relation types.
The total number of relation types is 42.
The dataset is obtained from the LDC website.\footnote{\url{https://catalog.ldc.upenn.edu/LDC2018T24}}
We compute the reported results using our code based on that of \newcite{Zhang2019}.

\subsection{CoNLL-2003}

The CoNLL-2003 dataset comprises training, development, and test sets, containing 14,987, 3,466, and 3,684 sentences, respectively.
Each sentence contains annotations of four entity types, namely \textit{person}, \textit{location}, \textit{organization}, and \textit{miscellaneous}.
The dataset is downloaded from the relevant website.\footnote{\url{https://www.clips.uantwerpen.be/conll2003/ner}}
The reported results are computed using the \texttt{conlleval} script obtained from the website.

\subsection{ReCoRD}

The ReCoRD dataset consists of 100,730 training, 10,000 development, and 10,000 test questions created based on 80,121 unique news articles.
The dataset is obtained from the relevant website.\footnote{\url{https://sheng-z.github.io/ReCoRD-explorer}}
We compute the performance on the development set using the official evaluation script downloaded from the website.
Performance on the test set is obtained by submitting our model to the leaderboard.

\subsection{SQuAD 1.1}

The SQuAD 1.1 dataset contains 87,599 training, 10,570 development, and 9,533 test questions created based on 536 Wikipedia articles.
The dataset is downloaded from the relevant website.\footnote{\url{https://rajpurkar.github.io/SQuAD-explorer}}
We compute performance on the development set using the official evaluation script downloaded from the website.
Performance on the test set is obtained by submitting our model to the leaderboard.

\section{Adding Entity Annotations to SQuAD dataset}
\label{sec:squad-annotations}

For each question--passage pair in the SQuAD dataset, we first create a mapping from the entity names (e.g., ``U.S.'') to their referent Wikipedia entities (e.g., \textit{United States}) using the entity hyperlinks on the source Wikipedia page of the passage.
We then perform simple string matching to extract all entity names in the question and the passage, and treat all matched entity names as entity annotations for their referent entities.
We ignore an entity name if the name refers to multiple entities on the page.
Further, to reduce noise, we also exclude an entity name if its \textit{link probability}, the probability that the name appears as a hyperlink in Wikipedia, is lower than 1\%.
We use the March 2016 version of Wikipedia to collect the entity hyperlinks and the link probabilities of the entity names.

\end{document}